# 3D Automatic Segmentation Method for Retinal Optical Coherence Tomography Volume Data Using Boundary Surface Enhancement

**Yankui Sun**[1]*
Department of Computer Science and Technology
Tsinghua University
Beijing 100084, China
syk@mail.tsinghua.edu.cn
*Corresponding author

**Tian Zhang**[1]
Department of Computer Science and Technology
Tsinghua University
Beijing 100084, China
kasperlzhang@gmail.com

**Yue Zhao**[1]
Department of Computer Science and Technology
Tsinghua University
Beijing 100084, China
zhaoyue_82@163.com

Yufan He[2]
Department of Electronic Engineering
Tsinghua University
Beijing 100084, China
heyufan1995@qq.com

**The running title:** 3D Segmentation for Retinal OCT Data

**Abstract—** With the introduction of spectral-domain optical coherence tomography (SDOCT), much larger image datasets are routinely acquired compared to what was possible using the previous generation of time-domain OCT. Thus, there is a critical need for the development of 3D segmentation methods for processing these data. We present here a novel 3D automatic segmentation method for retinal OCT volume data. Briefly, to segment a boundary surface, two OCT volume datasets are obtained by using a 3D smoothing filter and a 3D differential filter. Their linear combination is then calculated to generate new volume data with an enhanced boundary surface, where pixel intensity, boundary position information, and intensity changes on both sides of the boundary surface are used simultaneously. Next, preliminary discrete boundary points are detected from the A-Scans of the volume data. Finally, surface smoothness constraints and a dynamic threshold are applied to obtain a smoothed boundary surface by correcting a small number of error points. Our method can extract retinal layer boundary surfaces sequentially with a decreasing search region of volume data. We performed automatic segmentation on eight human OCT volume datasets acquired from a commercial Spectralis OCT system, where each volume of data consisted of 97 OCT images with a resolution of $496 \times 512$; experimental results show that this method can accurately segment seven layer boundary surfaces in normal as well as some abnormal eyes.

**Keywords**: Optical coherence tomography; retinal OCT volume data; 3D automatic segmentation

## 1. Introduction

The retina is a complex organization composed of a transparent layer of tissue. Automatic segmentation algorithms that accurately detect the layer structures in frequency-domain OCT retinal images are critical for the efficient diagnosis of ocular diseases such as glaucoma, diabetic retinopathy, etc. Many OCT image segmentation methods have been developed to segment the retinal layer boundaries with varying levels of success. Fernandez et al. proposed a method that used a structure tensor combined with complex diffusion filtering to segment seven retinal layer boundaries[1]; Mujat *et al.* implemented a method to determine the thickness of the retinal nerve fiber layer (RNFL) from OCT images by segmenting two boundaries using anisotropic noise suppression and deformable splines[2]; Ishikawa *et al*. recognized retinal layer positions by peaks and valleys in an A-scan intensity profile by using a mean filter for de-speckling, which segmented five layer boundaries[3]; Chiu *et al.* presented a segmentation method that used graph theory and dynamic programming to segment seven retinal layers[4];This method was later extended for segmentation of mouse retinal layers[5], anterior eye images[6] and age-related macular degeneration(AMD) images[7]; Using a similar method, Yang *et al.* [8,9] utilized a more complex approach to calculate the weights map of graph-based method, using dual-scale gradient information and shortest path search techniques to segment intra-retinal boundaries in OCT images. Yazdanpanah *et al.* used an active contour approach for the segmentation of rodent retinas[10]; A two-step kernel-based optimization was proposed by Mishra *et al.* [11]. However, the methods in [10,11] was never tested on OCT datadets of human retinas. Itebeddine *et al.* proposed a global segmentation algorithm based on using active contours and Markov random fields to segment eight retinal layers[12].

While the aforementioned methods can be used to segment OCT volume data slice-by-slice, most of them require long processing times, and they do not use the correlation between slices well. Recently, 3D OCT retinal image segmentation techniques have been developed. Zawadzki and Fuller *et al.* proposed segmentation methods using a support vector machine (SVM) and machine learning [13,14], which could segment one layer once by manual interaction. Similar to Zawadzki *et al.*' method, a support vector machine was used to classify pixels in the OCT image but in a fully automated way[15]. In [16], random forest classifier was built to segment eight retinal layers in macular cube images acquired by OCT. The random forest classifier learns the boundary pixels between layers, producing an accurate probability map for each boundary, which is then processed to finalize the boundaries. Kajic *et al.* proposed a method that used a large training dataset obtained from manual segmentations by human operators as input to develop a statistical model to segment seven retinal layers [17]. Garvin *et al.* proposed a graph search-based three-dimensional OCT retinal image segmentation algorithm [18,19], which could segment five retinal layers, which was later extended to incorporate hard/soft constraints[20]. Lee used multi-scale 3D graph search for segmenting the optic nerve head[21]. Compared to these complex 3D imaging segmentation approaches, Fabritius *et al.* presented a fast segmentation method for segmenting the internal limiting membrane (ILM) and the retinal pigment epithelium (RPE) that was based on variations in pixel intensity [22];While the method is relatively fast, only two boundaries are detected.

    To develop a more practical 3D segmentation technique such as computation efficiency and robust to blood vessel shadow and noise, here we propose a new 3D segmentation method for segmenting retinal layer boundaries from OCT volume data using boundary surface enhancement, which is relative simple, efficient, robust to

shadows and noise, and can segment seven boundary surfaces. To segment a boundary surface, two OCT volume datasets are obtained by using a 3D smoothing filter and a 3D differential filter. Their weighted sum is then calculated to generate new volume data with an enhanced boundary surface, where the pixel intensity, boundary position information, and intensity changes on both sides of the boundary surface are used simultaneously. Then, preliminary discrete boundary points are detected from the A-Scans of the volume data. Finally, surface smoothness constraints and a dynamic threshold are applied to obtain a smoothed boundary surface by correcting a small number of error points. Our methods can extract retinal boundary surfaces sequentially within a decreasing region of volume data. The key idea is to use pixel position information, gradient information and intensity information simultaneously to enhance the boundary surface to be detected so that preliminary discrete boundary points can be detected more correctly and error points can be eliminated more easily.

This paper is organized as follows: Section 2 gives a description of our generalized layer segmentation algorithm, which is fundamental for segmenting all of the layer boundary surfaces; Section 3 demonstrates how to segment seven retinal layer boundary surfaces in detail; experimental results and analysis are given in Section 4; and conclusions are made in Section 5.

## 2．A Generalized Layer Segmentation Algorithm

An image acquired from a commercial Spectralis OCT device (Heidelberg Engineering, Heidelberg, Germany ) is shown in Fig. 1, where the left panel shows the scanning position in the retinal tissue and the right panel shows the corresponding SDOCT image. Fig. 2 illustrates a volume dataset made up of a sequence of SDOCT

B-scans in the *y* direction. Every B-scan is composed of A-scans in the *x* direction. Each A-scan has a depth coordinate *z*, which increases going from top to bottom in the image.

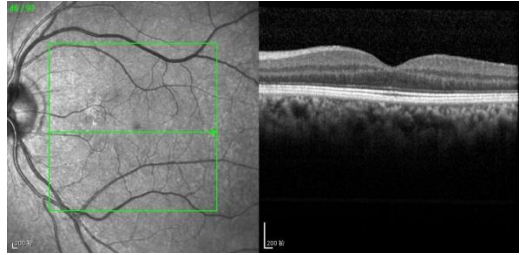

**Fig.** 1. SDOCT image

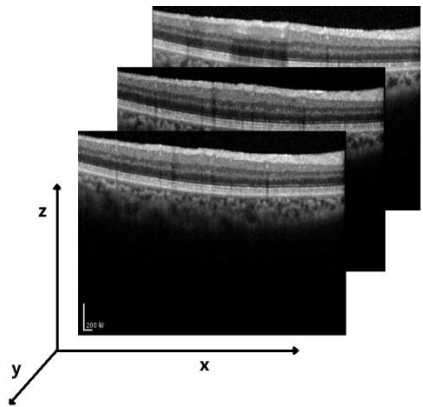

**Fig**. 2. Illustration of a volume dataset.

A retinal OCT image consists of layer structures (Fig. 3), where the intensity varies in the layers due to differences in the reflection properties of the retinal tissue. Moreover, layer boundaries have various orientations such as the Vitreous-ILM layer boundary, which exhibits a dark layer above a light layer.

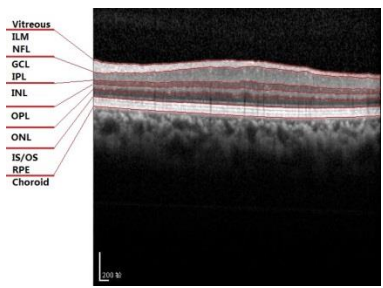

**Fig**. 3. Retinal OCT image with layer structures.

This section proposes a method for segmenting the layer structures of the retina. The basic idea of this method is to use the characteristics of the boundary of interest to design a 3D operator and apply it to the original volume data so that the pixel value on the desired boundary in the new volume data is likely to be the maximum value in its A-scan. This approach makes the new volume data a better indicator of the desired boundaries. The algorithm consists of three steps: denoising, extracting boundary points and correcting error points. Obviously, the key problem is to determine how to identify the correct discrete boundary points. Our algorithm tries to achieve accurate boundary point detection by enhancing the boundary of interest. The core steps in our basic retinal layer boundary surface segmentation (BRLBSS) algorithm are below.

**Step 1**: Denoising

A three-dimensional average filter with size of $K_1 \times K_2 \times K_3$ is applied to the original retina OCT volume data to obtain a smoothed volume data, $S$, where $K_1, K_2$ and $K_3$ stand for the filter window width in the *x*, *y*, and *z* directions, respectively, and are all positive odd numbers.

**Step 2:** Extract the boundary points

1) *Enhance the boundary with gradient information*

A three-dimensional differential filter with a size of $M_1 \times M_2 \times M_3$ is applied to the original retina OCT volume data to obtain a differential volume data, *D*, where $M_1, M_2$ and $M_3$ ($M_3 \neq 1$) are all positive odd numbers, similar to $K_1, K_2$ and $K_3$.

For an original volume data, $V$, and one boundary of interest to be segmented, a pixel $v_{i_0, j_0, k_0}$ in $V$, is taken as a center point, and a cuboid with size $M_1 \times M_2 \times M_3$ is constructed using the pixel and its adjacent pixels, $v_{i,j,k}$, where *i, j, and k* are the *x, y, and z* coordinates of each pixel. The differential filter is sensitive to the boundary orientation.

For the RPE-Choroid, OPL-ONL, IPL-INL and NFL-GCL boundaries,

$$f_{i,j,k} = \begin{cases} 1, k < k_0 \\ 0, k = k_0 \\ -1, k > k_0 \end{cases} \quad (1)$$

For the Vitreous-ILM, ONL-IS/OS and INL-OPL boundaries,

$$f_{i,j,k} = \begin{cases} -1, k < k_0 \\ 0, k = k_0 \\ 1, k > k_0 \end{cases} \quad (2)$$

Then, the differential filter is used to obtain a differential volume data, *D*. Specifically, for every cuboid in *V*, the intensity sum of the pixels above/below the cubiod center is subtracted from that of the pixels below/above the cuboid center, and the result is averaged for $M_1 \cdot M_2 \cdot (M_3 - 1)$ to obtain the intensity of the cuboid center.

   2) *Enhance the boundary by position, gradient and intensity*

The depth, gradient and intensity information for the boundary are used to generate a boundary-enhanced volume data, *I*, to further highlight the boundary of interest. *I* is calculated as follows:

$$I_{i,j,k} = w_1 D_{i,j,k} + w_2 S_{i,j,k} \quad (3)$$

where *i, j, and k* are the *x, y,* and *z* coordinates of each pixel, and $w_1, w_2$ are non-negative real numbers that are directly proportional to the depth coordinate, *k*.

   The intensity of the desired boundary in *I* is likely to be the maximum value in its

A-scan.

3) *Extract the boundary surface points*

In each A-scan of $I$, the point with the highest pixel intensity or the first peak (up to down or down to up, depending on the direction of the boundary of interest) is taken as the preliminary location of the desired boundary. Because the highest pixel intensity value is likely to lie on the desired boundary positions, the results obtained by this approach generally produce an accurate estimate of the location of the boundary.

**Step 3**: Correct the error points

The depth positions with large errors are corrected using surface smoothness constraints, which require that the difference between the z coordinates of adjacent pixels is small. In step 2, the boundary surface preliminary positions (z coordinates) make up a depth information matrix, $A$. Given a weighted matrix, $W_1$, for a depth element, $p$, in $A$, the absolute value of the difference between $p$ and the weighted average of its adjacent entries is called the *error distance* (ED) of the element (associated with $W_1$). For a threshold $T$, if the error distance of the element $p$ is larger than the threshold, then $p$ is considered to be an *error point* (associated with the threshold $T$). Given a weighted matrix $W_2$, if an element $p$ in $A$ is an error point, the weighted average of its adjacent entries with $W_2$ is taken as its *correcting value (associated with $W_2$)*. The matrix $A$ can be smoothed through iteration as follows:

Choose a number of iterations, $N$. For the $i$th iteration, pick a threshold, $T_i$. For each entry in $A$, calculate its error distance. If an entry is an error point, then it is replaced by its corresponding *correcting value*. If there exists at least one error point and the iteration $i$ is not equal to $N$, then advance to the next iteration until either there are no error points or *N* is reached.

The final depth information in *A* constitutes the desired boundary surface.

## 3. Implementation of the Algorithm for Segmenting Seven Retinal Layer Boundary Surfaces

This section details the implementation of the segmentation algorithm in Section 2 that automatically segments seven retinal boundary surfaces in SDOCT volume data. Fig. 4 shows a full schematic of this algorithm, and the following subsections discuss each of the outlined steps.

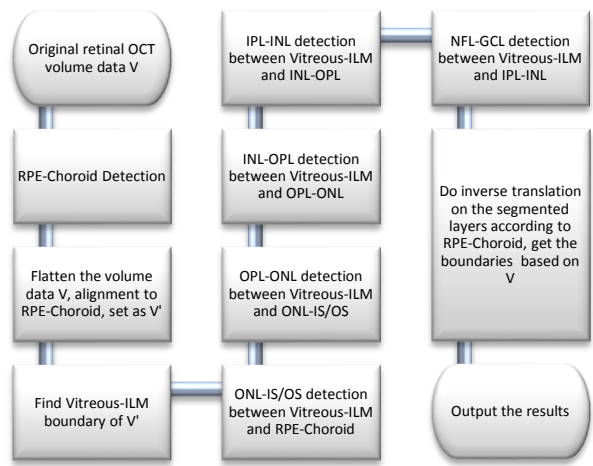

**Fig**. 4. Seven retinal layer boundary surface segmentation algorithm schematic for SDOCT volume data.

*3.1 RPE-Choroid boundary surface detection*

Based on prior knowledge, the RPE layer is one of the most hyper-reflective layers within a retinal SDOCT image. Thus, the RPE-Choroid boundary surface detection is processed first and its implementation is performed according to the layer boundary detection algorithm in Section 2. In our implementation, $K_j = M_j = 7, j = 1, 2, 3$ and $w_1 = w_2 = k$. The two weighted matrices are not unique. We take them as:

$$W_1 = \frac{1}{138}\begin{bmatrix} 0 & 1 & 1 & 1 & 1 & 1 & 0 \\ 1 & 2 & 2 & 2 & 2 & 2 & 1 \\ 2 & 4 & 4 & 4 & 4 & 4 & 2 \\ 4 & 8 & 16 & -138 & 16 & 8 & 4 \\ 2 & 4 & 4 & 4 & 4 & 4 & 2 \\ 1 & 2 & 2 & 2 & 2 & 2 & 1 \\ 0 & 1 & 1 & 1 & 1 & 1 & 0 \end{bmatrix}, \quad W_2 = \frac{1}{138}\begin{bmatrix} 0 & 1 & 1 & 1 & 1 & 1 \\ 1 & 2 & 2 & 2 & 2 & 2 \\ 2 & 4 & 4 & 4 & 4 & 4 \\ 4 & 8 & 16 & 0 & 16 & 8 \\ 2 & 4 & 4 & 4 & 4 & 4 & 2 \\ 1 & 2 & 2 & 2 & 2 & 2 & 1 \\ 0 & 1 & 1 & 1 & 1 & 1 & 0 \end{bmatrix}.$$

For smoothing, the number of iterations also should be determined reasonably. According to our various experiments on the test datasets, we choose the number of iterations $N=25$.. For the first 20 iterations, a dynamic threshold is used according to $T_i = count \times i / 2000$, where *count* is the element number in matrix *A*, *i* is the iteration number and 2000 is an empirical value. In the last 5 iterations, the threshold is fixed at 1.

Our algorithm can process retinal SDOCT images with prominent vessels. First, the RPE-Choroid boundary in every B-scan is relatively flat. Thus, choosing filters with relatively larger windows can smooth the original data more to reduce the influence of vessel containing regions. Additionally, a dynamic threshold is used to quickly and accurately correct outlier points. Fig. 5 shows the existence of outlier points before smoothing.

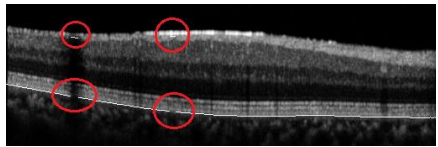

**Fig**. 5 Great outlier points exist before smoothness

Fig. 6 shows the processed result for Fig. 5 after five iterations using a dynamic threshold. It can be seen that the outlier points approach correct points. Fig. 7 shows the final smoothed result after 20 iterations.

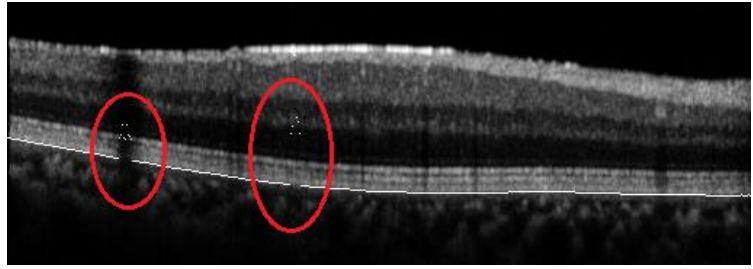

**Fig**. 6. The processed result of Fig. 5 after 5 iterations by using dynamic threshold.

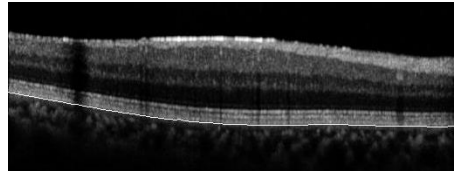

**Fig**. 7. The final smoothed result after 20 iterations.

Some retinal layers may have large curvature in the SDOCT images, such as the fovea region. To correct for curvature, we flatten the images to enhance their adaptation for the segmentation algorithm. In this process, the image below the RPE-Choroid boundary is ignored. Fig. 8 demonstrates retinal flattening, where Fig. 8b is the flattened version of the original image, Fig. 8a.

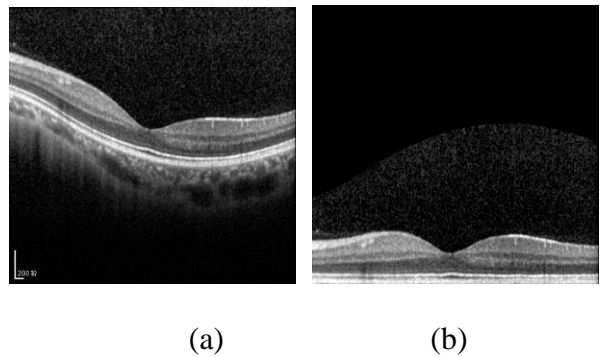

(a)         (b)

**Fig**. 8 Image flattening. (a) The original retinal SDOCT image. (b) The flattened image without the region below the RPE-Choroid boundary

*3.2 Vitreous–ILM boundary surface segmentation*

Vitreous–ILM boundary surface detection is processed on the flattened volume data of the original volume data *V* by ignoring the pixels below the RPE-Choroid boundary surface. For a low noise retinal OCT image where only background noise exists above the ILM, similar to the RPE-Choroid boundary surface detection, the BRLBSS algorithm in Section 2 can be used to obtain an accurate segmentation result. However, because the Vitreous–ILM boundary may have a large curvature in the fovea region, the choice of the related parameters should be considered carefully. For example, relatively narrow three-dimensional filters in the *x* direction (Fig. 2) should be used for the fovea region.

**Vitreous–ILM boundary surface segmentation (VI_BSS) algorithm:**

**Step 1**: Denoising

For the flattened volume data $V'$, a three-dimensional average filter with window size $6\times6\times6$ is used to produce a volume data denoted $M$. Then, a threshold filter is applied to *M*. In our experiment, the threshold value was 30. This filter procedure was iterated several times to obtain a smoothed volume data, *S*.

**Step 2:** Extract boundary points

*1) Enhance the boundary with gradient information*

A differential filter in the *z* direction is applied to the volume data *S* using a cuboid with a size of $1\times1\times11$. The new volume data is denoted by *D*.

*2) Extract discrete boundary points*

In every A-Scan of the volume data *D*, the first peak point is determined by moving from up to down. These points constitute the preliminary Vitreous–ILM boundary surface.

**Step 3**: Correct error points

This step is similar to the RPE-Choroid boundary surface smoothness scheme. The main difference is the smaller size of the weighted matrices that are used to adapt the large curvature features of ILM layer in the fovea region. The weighted matrices are:

$$W_1 = \frac{1}{64}\begin{bmatrix} 0 & 1 & 2 & 1 & 0 \\ 1 & 4 & 8 & 4 & 1 \\ 2 & 8 & -64 & 8 & 2 \\ 1 & 4 & 8 & 4 & 1 \\ 0 & 1 & 2 & 1 & 0 \end{bmatrix} \quad W_2 = \frac{1}{64}\begin{bmatrix} 0 & 1 & 2 & 1 & 0 \\ 1 & 4 & 8 & 4 & 1 \\ 2 & 8 & 0 & 8 & 2 \\ 1 & 4 & 8 & 4 & 1 \\ 0 & 1 & 2 & 1 & 0 \end{bmatrix}.$$

In the last 5 iterations, the size of weighted matrices is reduced further. They are:

$$W_1 = \frac{1}{12}\begin{bmatrix} 1 & 2 & 1 \\ 2 & -12 & 2 \\ 1 & 2 & 2 \end{bmatrix} \quad W_2 = \frac{1}{12}\begin{bmatrix} 1 & 2 & 1 \\ 2 & 0 & 2 \\ 1 & 2 & 2 \end{bmatrix}.$$

The VI_BSS algorithm is effective for OCT images with vessel shadows and/or a fovea region. However, it may fail for clinical OCT images with significant noise above the ILM (Fig. 9). To adapt this case, an improved algorithm is proposed.

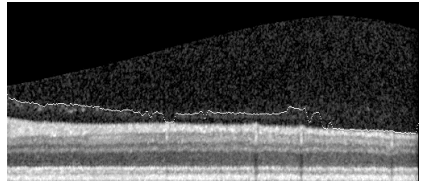

**Fig**. 9. Incorrect segmentation because of serious noise above ILM.

**Vitreous–ILM preliminary boundary surface segmentation (VI_PBSS) algorithm:**

**Step 1**: Denoising

Gray-scale morphological corrosion with a ball structure of radius $r$ is performed on the smoothed volume data $S$ obtained from the VI_BSS algorithm. Then, the average and threshold filter processing is repeated, as done in Step 1 of the VI_BSS algorithm. We denote the denoised volume data with $SE$. In our implementation, $r = 5$ was used.

**Step 2:** Extract the preliminary boundary points

  *1) Enhance the boundary with gradient information.*

A differential filter in the $z$ direction is applied to the volume data $SE$ using a cuboid with a size of 1x1x11. The new volume data is denoted by $D$.

2) *Extract discrete boundary points*

In every A-Scan of the volume data $D$, determine the first peak point moving from up to down. To compensate for the excessive erosion, $\lceil r/2 \rceil$ pixels are subtracted from the depth coordinate, and these points constitute the preliminary Vitreous–ILM boundary surface points.

The VI_PBSS algorithm works well for OCT images with considerable noise above the ILM, but it may fail for clinical OCT images with prominent vessels (Fig. 10) due to excessive erosion.

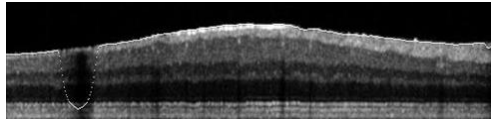

**Fig**. 10. Incorrect segmentation results because of excessive erosion in a vessel or shadow region.

Next, we merge the boundary points obtained from the VI_BSS and VI_PBSS algorithms to obtain the final segmentation result. The basic strategy is to replace the discontinuous boundary points obtained from the VI_PBSS algorithm with the corresponding boundary points obtained from the VI_BSS algorithm.

Let $A$ and $B$ be the boundary depth matrices obtained from the VI_PBSS and VI_BSS algorithms, respectively. Matrix $C$ is the average filter result with a window size of $n \times n$ on the boundary depth matrix $A$. In our implementation, $n = 11$ was used. Matrix $E$ is the merged boundary depth matrix. If $|a_{ij} - c_{ij}| < |b_{ij} - c_{ij}|$, then $e_{ij} = a_{ij}$; otherwise $e_{ij} = b_{ij}$. Finally, the depth matrix $E$ is further smoothed according to step 3 used in the VI_BSS algorithm to obtain the Vitreous–ILM boundary surface segmentation result. Fig. 11 shows the segmentation results with the improved algorithm.

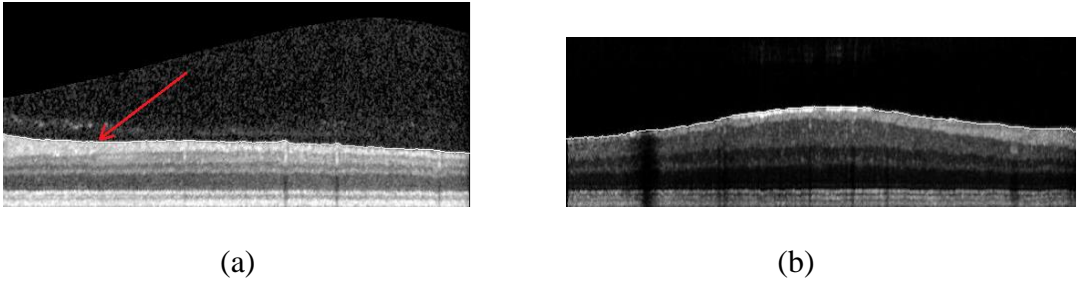

(a)                                      (b)

**Fig**. 11. Segmentation results with the improved algorithm. (a) Image with considerable noise; (b) Image with a large shadow.

Once the RPE-Choroid and Vitreous-ILM boundary surfaces are determined, the middle boundary surfaces between them, ONL-IS/OS, OPL-ONL, INL-OPL, IPL-INL, and NFL-GCL, can be segmented sequentially moving from down to up in a smaller and smaller search space.

*3.3 ONL-IS/OS boundary surface segmentation*

ONL-IS/OS boundary surface segmentation can be done with the flattened volume data *V* in between the RPE-Choroid and Vitreous-ILM boundary surfaces, where pixel intensities are set to 0 below the RPE-Choroid boundary and to 255 above the Vitreous-ILM boundary. We apply the three-dimensional differential filter in Eq. (2) with a size of $3\times3\times11$ on the flattened volume data to obtain new volume data, *D*. In Eq. (3), $w_1 = k$ and $w_2 = 0$ are used to obtain an enhanced volume data, *I*. In every A-scan of data *I*, the point with the largest pixel value is taken as the preliminary location of the desired boundary. In principle, the ONL-IS/OS boundary surface can be obtained from step 3 in the BRLBSS algorithm. However, this algorithm may not work well for some OCT images with dark spots in the fovea region and near RPE, as shown in Fig. 12.

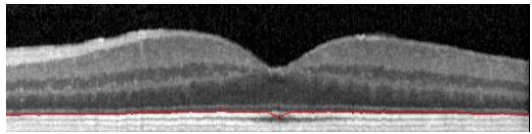

**Fig**. 12. Incorrect segmentation result in the fovea region and near the RPE because of dark spots.

To adapt our algorithm to this case, another error correction procedure can be added before step 3 in the BRLBSS algorithm. Because the IS/OS layer is relatively flat, a third order polynomial can be applied to rule out some of the error points. This procedure works as follows:

1) For the preliminary boundary points of each B-scan in *I*, polynomial least squares estimates is performed to obtain a fitted polynomial

$$z = a_0 + a_1 x + a_2 x^2 + a_3 x^3.$$

2) The depth value of every boundary point is substituted into the above expression to obtain a polynomial fitted value. If the depth value is outside of the confidence interval for the estimated value associated with a probability of 0.98, the boundary point is considered to be a noise point and is eliminated. The remaining credible boundary points are used to repeat the polynomial least squares estimate and obtain another fitted polynomial

$$z_1 = a_4 + a_5 x + a_6 x^2 + a_7 x^3.$$

3) The *x* coordinate of every noise point is substituted into the polynomial $z_1$, and the polynomial fitted value is taken as a depth estimate for the noise boundary point.

With the last obtained preliminary ONL-IS/OS boundary points, the error points are corrected using the same algorithm that is used for the RPE-Choroid smoothing.

The ONL-IS/OS boundary surface can be segmented correctly by using a polynomial fitting to eliminate the error points (Fig. 13).

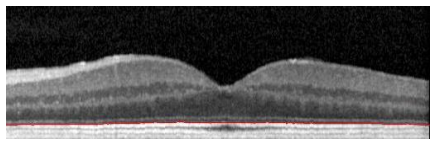

**Fig**. 13. ONL-IS/OS boundary surface segmentation result obtained by using polynomial fitting to eliminate the error points.

*3.4 OPL-ONL, NFL-GCL, IPL-INL, INL-OPL boundary surface segmentations*

To segment the OPL-ONL boundary surface, we further shrink the search space in between the ONL-IS/OS and Vitreous-ILM layer boundaries. The basic steps are similar to the steps used in the ONL-IS/OS boundary segmentation. The differences are as follows:

1) The OPL-ONL boundary does not appear dark spots like the ONL-IS/OS boundary, and it may have a large curvature when the fovea region lies in the image. As a result, the step using a third-order polynomial to eliminate some error points is not used here.

2) The 3D differential filter in Eq. (1) with size of $7\times15\times15$ on the flattened volume data is used.

The other boundary detection algorithms for the NFL-GCL, IPL-INL, and INL-OPL are similar to that used for OPL-ONL boundary detection. When the Vitreous–ILM, ONL-IS/OS, OPL-ONL, NFL-GCL, IPL-INL, and INL-OPL segmentations are all completed with the flattened volume data, an inverse translation on the segmented layers is performed according to the RPE-Choroid boundary to obtain the boundaries based on the original volume data *V*.

**4．Experimental results and analysis**

To determine the segmentation accuracy of our algorithms for SDOCT volume data, we performed segmentation on eight OCT volume datasets acquired from a commercial Spectralis OCT device, using Matlab R2012b (The Mathworks, Inc., Natick, MA, USA). Each volume of data consisted of 97 OCT images with a resolution of $496\times512$ and saved in AVI video format. The OCT images were first read from the AVI file. In our experiments, manual segmentation and the automatic segmentation made by our algorithm are done and their experimental results are compared. We manually delineated an image frame using the internally developed matlab GUI. We choose ten frames randomly from each OCT volume datasets, and delineated 7 layers manually on each frame. The manual delineations were performed by clicking on approximately 20–50 points along each layer border followed by interpolation between the points using a cubic B-spline. And we calculated the absolute and signed errors for each layer and averaged them. Segmentation results of some frames are illustrated in Fig. 14, where the manually delineated boundary and the automatically segmented boundary are plotted in the red line and green line respectively. When one color covers the other, that means our algorithm fits well with the manual method at that position. The pictures show that our algorithm is quite accurate no matter whether the image has a fovea or an incline. Table 1 shows that the average absolute error compared to manual delineation is 4.05μm, and the thickness of the thinnest layer in OCT volume is about 25 μm (Table 2). Compared to the thickness, the absolute error is acceptable.

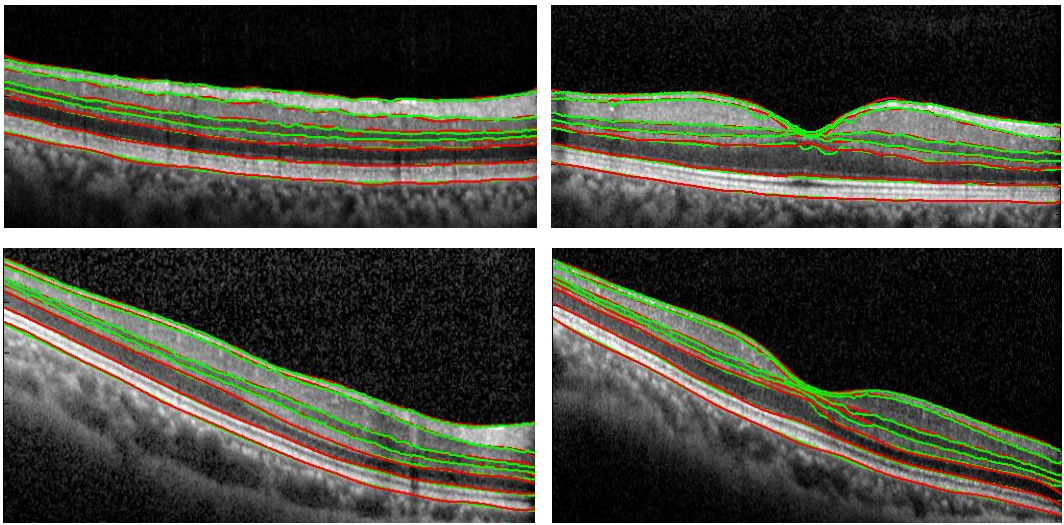

**Fig**. 14 Segmentation results of some frames

Table 1. Mean Absolute and Signed Errors(and Standard Deviations) in μm

| Boundary | Absolute Errors(μm) | Signed Errors(μm) |
| --- | --- | --- |
| Vitreous -ILM | 4.0198(3.2341) | 2.4100(4.1360) |
| NFL-IPL | 4.7745(4.9102) | -1.1015(6.3089) |
| IPL-INL | 4.5762(4.2947) | -0.3850(5.6009) |
| INL-OPL | 4.9414(4.2379) | -3.1104(5.3200) |
| OPL-ONL | 3.9624(3.1920) | -1.0156(4.3351) |
| ONL-IS/OS | 2.8599(2.0804) | 1.0526(2.9858) |
| RPE-Choroid | 3.2169(2.3709) | 0.0844(3.4541) |
| Overall | 4.0502(3.4743) | -0.2951(4.5916) |

To further examine our algorithm, we calculated the thickness of the OCT layer from manual delineated data and auto-segmentation data. The experimental results are given in Table 2. The results listed in Table 1 and Table 2 are very similar in essential, which shows that our methods are rather practical for substituting the manual segmentation method.

Table 2. Manual Thickness and Auto-segment thickness in μm

| Layer | Manual Thickness(μm) | Auto-segment Thickness (μm) |
| --- | --- | --- |
| NFL | 42.8851 | 46.3300 |
| GCL | 61.4994 | 60.8878 |
| INL | 25.0831 | 27.6770 |
| OPL | 29.8844 | 27.8779 |
| ONL | 65.7441 | 63.6382 |
| RPE | 60.6234 | 61.6538 |

To further analyze the stability of our algorithm, we calculated the mean absolute errors for each OCT volume (Fig.15). It shows that the overall error for each subject is quite stable, but the segmentation of the layer INL-OPL is quite conditional on the quality of the data. Anyway, according to Table 1 and Table 2, these errors could be acceptable comparing to the thickness of the layer.

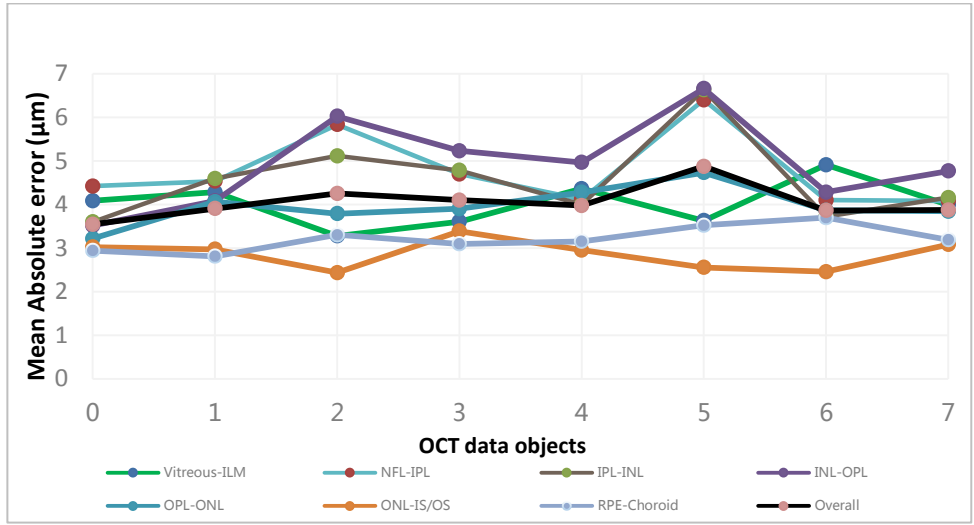

**Fig. 15**.   Mean Absolute error of eight different OCT data objects

Fig. 16 presents some experimental results for various cases. It can be seen that our automatic algorithm correctly segmented seven retinal layer boundary surfaces for retinal OCT images with large curvature, severe noise above the ILM, fovea, prominent vessels, or large shadows because of diseases. The average computation time was 3.5 seconds per frame (Intel Core I7 CPU at 3.0GHz, and 8GB RAM), and the program can be optimized to reduce the time further. Therefore, our algorithm is simple and fast.

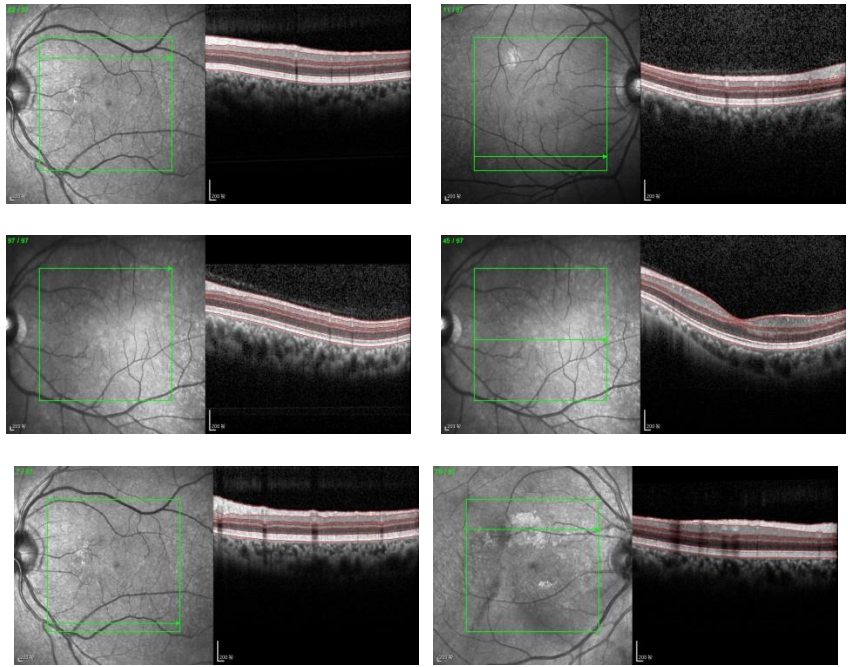

**Fig** .16. Some experimental results from various imaging cases.

Once the retinal layer segmentations have been performed successfully, every layer boundary surface can be visualized, and a thickness map between any two boundaries can be generated, which can be useful for disease diagnosis. An example visualization of the Vitreous-ILM and RPE-Choroid boundary surfaces is shown in Fig. 17, and the whole retinal thickness map determined from them is shown in Fig. 18.

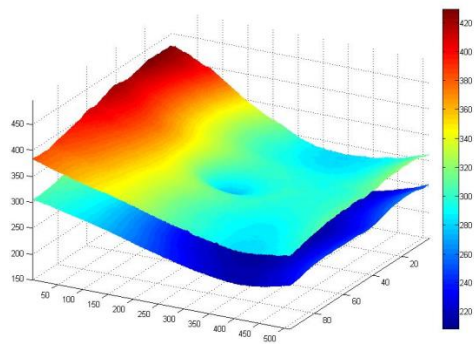

**Fig**. 17. Visualization of the Vitreous-ILM and RPE-Choroid boundary surfaces.

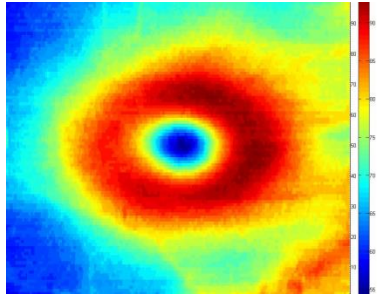

**Fig**.18. Retinal thickness map determined from the Vitreous-ILM and RPE-Choroid boundary surfaces.

Our method is inspired by the work in [22], but it also differs from it greatly. In principle, the method of Fabritius et al. only uses the pixel intensity of A-scans to detect the preliminary boundary points, and then boundary position information is used to eliminate the error points iteratively. As a result, their method is sensitive to noise and is invalid for images with severe noise. Our method uses pixel position information, gradient information and intensity information simultaneously to extract the boundary points so that the number of large error points is greatly reduced. Thus, the error point correction method used in our algorithm is completely different. Moreover, our algorithm can segment seven retinal boundary surfaces rather than just two. Compared to the complex 3D graph search-based methods in [18,19], our method is simple in principle and every efficient, and it does not depend on more prior knowledge and training dataset like the method in [20]. Additionally, our algorithm can deal with retinal OCT images with considerable noise above the ILM, and it does not need to detect prominent vessels or the fovea region explicitly. In our algorithm, the most three important boundary surfaces (i.e., RPE-Choroid, Vitreous-ILM and ONL-IS/OS) are segmented first, which can be used in improving some other algorithms such as the ones developed in [4,18,19] to shrink their search space at first, as those methods require a considerable amount of time to segment these boundaries because of the large search space.

As is the case for most of the current OCT image segmentation algorithms, our algorithm processes healthy or slightly abnormal retinal OCT images well but fails to accurately process retinal OCT images with serious diseases (Fig. 19).

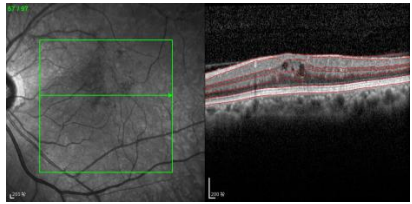

**Fig**. 19. Segmentation result for a retinal OCT image with serious abnormalities.

## 5．Conclusions

We propose a novel 3D segmentation method for retinal OCT volume data that uses pixel intensity, boundary position information, and intensity changes on both sides of the layer borders simultaneously. The method designs a specific 3D differential operator for the processing boundary to enhance the border, it conducts a three-dimensional smoothing procedure to denoise the volume data, and it further utilizes the boundary position to produce an enhanced boundary volume data that serves as a better indicator for identifying the desired boundaries. Our method can segment seven boundary surfaces, and it is automatic, efficient and practical.

**Acknowledgments**

This research was supported by the National High Technology Research and Development Program of China ("863" Program) under Grant No. 2013AA013702 and the National Natural Science Foundation of China (No. 60971006).